\documentclass[letterpaper, 10 pt, conference]{ieeeconf}  % Comment this line out if you need a4paper

\IEEEoverridecommandlockouts                              % This command is only needed if 
\usepackage{graphics} % for pdf, bitmapped graphics files
\usepackage{epsfig} % for postscript graphics files
\usepackage{mathptmx} % assumes new font selection scheme installed
\usepackage{times} % assumes new font selection scheme installed
\usepackage{amsmath} % assumes amsmath package installed
\usepackage{amssymb}  % assumes amsmath package installed
\usepackage{xspace} % for smart spacing after commands
\usepackage{booktabs}
\usepackage{xcolor}
\usepackage{cite}
\usepackage{url}

\newcommand{\core}{PolyMorphWBC\xspace}

\title{\LARGE \bf
\core: Whole-Body Control from End-Effector Trajectories for Quadrupedal Mobile Manipulation across Arm Morphologies with Extended Reach
}

\author{Heng Li, Yang Li, Yifan Duan, Ren Fang, Matao Duan, Wenhao Yu, Lu Zhang*, \\ Jianmin Ji and Yanyong Zhang*
\thanks{* The corresponding author.}
}%

\begin{document}

\maketitle
\thispagestyle{empty}
\pagestyle{empty}

\begin{abstract}

Quadrupedal mobile manipulators can extend their manipulation workspace by coordinating the mounted arm with the mobile base. 
% At the task level, short-horizon end-effector (EE) trajectories provide a task-centric interface for visuomotor and vision-language-action (VLA) policies. 
At the task level, short-horizon end-effector (EE) trajectories provide an interface for specifying the desired manipulation motion without prescribing how it should be realized by the arm and base.
% However, executing such trajectories still requires a whole-body controller to determine morphology-dependent base motion and compensate for arm--induced disturbances. 
% However, executing such trajectories still requires a whole-body controller to determine when and how the base should move with different arm morphologies.
However, executing such trajectories remains challenging because the controller must determine when and how the base should move, and this decision depends on the mounted-arm morphology.
% However, executing such trajectories still requires a whole-body controller to determine when and how the base should move, particularly as trajectories approach or exceed fixed-base reach and as the required coordination varies across arm morphologies.
We therefore present \core, a hierarchical controller that takes an EE trajectory as the sole task-level motion command and generates coordinated arm--base motions across four families of 6-DoF mounted arms, with extended-reach trajectories beyond the initial fixed-base reach.
% We call a trajectory \emph{extended-reach} if at least one desired EE pose that is infeasible for the arm under joint-limit and collision constraints with the base fixed at its initial pose. 
Within \core, a trajectory-conditioned coordinator determines the base motion from previewed EE targets, while arm-specific Inverse Kinematics (IK) tracks the current target, and a frozen locomotion backend executes the resulting motion. 
Cross-arm reuse is enabled by a Unified Robot Description Format (URDF)-derived morphology conditioning, together with online residual-wrench feedback.
We train \core on procedurally generated arms and structured trajectories within and beyond fixed-base reach. 
In IsaacLab, it achieves 93.95\% completion with 2.30\,cm position error and 4.81$^\circ$ orientation error on held-out trajectories and arm morphologies. 
% In simulations, it achieves 93.95\% completion with 2.30\,cm position error and 4.81$^\circ$ orientation error on held-out trajectories and arm morphologies, while retaining 91.56--94.83\% completion on extended-reach subsets.
In MuJoCo evaluation against UMI-on-Legs, \core reduces EE position error by 69--72\% and orientation error by 91--97\% across four within- and extended-reach trajectories.
% On a physical Go2--ARX5 system, we demonstrate language-conditioned pick-and-place, including extended-reach manipulation, using EE trajectories from $\pi_0$ without VLA-generated base commands.
On a physical Go2--ARX5 system, we demonstrate language-conditioned extended-reach pick-and-place using EE trajectories from $\pi_0$ without vision-language-action (VLA)-generated base commands.
% These results support a common trajectory-to-WBC interface across arm morphologies and beyond fixed-base reach. 
% Our code will be released upon publication.

\end{abstract}    
\section{Introduction}
\label{sec:introduction}

\begin{figure}[t]
    \centering
    \includegraphics[width=\columnwidth]{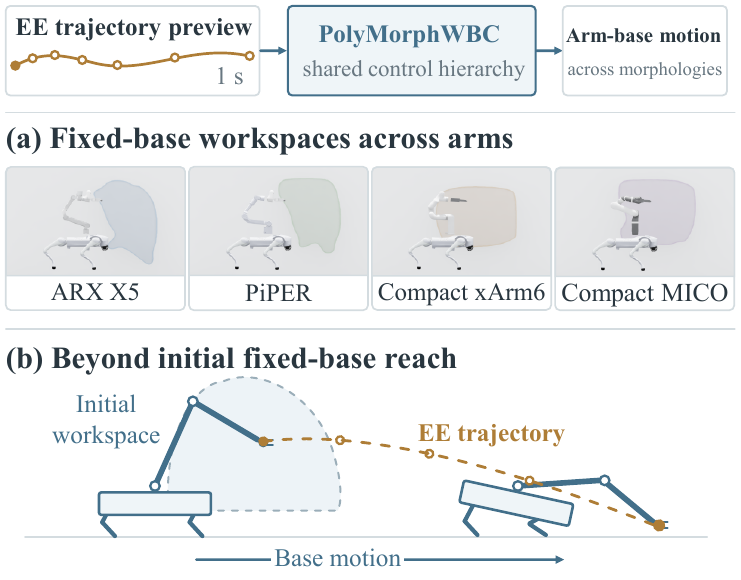}
    \caption{\textbf{Cross-arm reuse and extended reach with \core.}
    Top: a shared control hierarchy maps short-horizon EE trajectories to morphology-dependent arm--base motion across mounted-arm morphologies.
    (a) Representative arms exhibit different fixed-base workspaces. %, shown as translucent shaded regions.
    (b) Coordinated base motion enables tracking of EE trajectories extending beyond the arm's initial fixed-base reach, without task-level base-motion commands. Panel (b) is schematic.}
    \label{fig:introduction}
    \vspace{-2.0em}
\end{figure}

Quadrupedal mobile manipulators combine legged mobility with the reach and dexterity of robotic arms, enabling manipulation beyond the arm's fixed-base workspace~\cite{legged_manipulation_review,mobile_manipulation_review}.
Exploiting this capability requires whole-body control (WBC) to coordinate the mounted arm with the mobile base while maintaining physical stability.
Existing approaches typically achieve such coordination through model-based optimization or learned control~\cite{alma,unified_mpc,deep_wbc,visual_wbc,roboduet}.
Despite substantial progress, task interfaces may still require explicit base-motion commands, while controllers are often tailored to a particular arm--base configuration.
These dependencies complicate integration with task-level policies and controller reuse across arm morphologies.

Meanwhile, recent visuomotor and VLA policies increasingly predict short action sequences~\cite{act,diffusion_policy,pi0}.
For mobile manipulation, a short-horizon EE trajectory provides a natural task-level interface: it specifies the intended manipulation motion without prescribing how motion should be distributed between the arm and the mobile base.
% To simplify task-policy integration, short-horizon EE trajectories provide a natural interface: they specify the intended manipulation motion without prescribing base motion.
% This interface aligns with short action sequences predicted by recent visuomotor and vision-language-action (VLA) policies~\cite{act,diffusion_policy,pi0}.
UMI-on-Legs~\cite{umi_on_legs} demonstrates EE-trajectory-driven posture adjustment and stepping beyond fixed-base reach, although spatial coverage remains a challenge for learned tracking controllers~\cite{cmp}. MLM~\cite{mlm} extends the EE-trajectory interface to multiple tasks.
However, these controllers are developed for fixed mounted-arm morphologies.
% This raises a broader controller-side question: can one control hierarchy execute EE trajectories both within and beyond fixed-base reach while generalizing across mounted-arm morphologies?
% This raises a broader controller-side question: can one control hierarchy retain the same EE-trajectory interface across arms with different fixed-base workspaces while also executing trajectories beyond initial fixed-base reach?
% This raises a broader controller-side question: can one control hierarchy execute EE trajectories across arms with different fixed-base workspaces, including trajectories that extend beyond initial fixed-base reach?
% This raises a broader controller-side question: can one shared control hierarchy retain a common EE-trajectory interface while generating the morphology-dependent arm--base motion required by arms with different fixed-base workspaces, including when trajectories extend beyond initial fixed-base reach?
Cross-arm reuse of the control hierarchy under the same EE-trajectory interface remains largely unexplored, especially for trajectories beyond initial fixed-base reach.
% Fig.~\ref{fig:introduction} illustrates these two aspects of the problem.
Fig.~\ref{fig:introduction} summarizes this trajectory-to-motion interface together with cross-arm workspace variation and reach extension through base motion.

Addressing this problem involves two coupled challenges.
First, extending manipulation reach requires the controller to anticipate when base motion is needed: keeping the base fixed may leave upcoming EE targets unreachable, whereas premature or excessive motion can degrade EE tracking.
% Second, this coordination is morphology dependent. Arm kinematics, joint limits, and collision geometry determine which targets require base motion, while mass distribution and arm motion affect the disturbances transmitted to the base.
Second, this coordination is morphology dependent. For the same EE trajectory, kinematic differences change the feasible arm motion and required base response, while mass distribution and realized arm motion change the disturbances transmitted to the base.
% These challenges are particularly pronounced on a compact quadrupedal base, where arm reachability, base motion, and stability must be coordinated under tight whole-body constraints.
We study these challenges on a compact Unitree Go2\footnote{Unitree Go2: \url{https://www.unitree.com/go2}} base, where arm reachability, base motion, and stability are tightly coupled.
A reusable hierarchy must therefore combine future task motion, arm morphology, and interaction feedback. %while retaining arm-specific kinematic handling.

In this work, we design and develop \core, a hierarchical controller that converts short-horizon EE trajectories into coordinated arm and base motion across mounted-arm morphologies.
The EE trajectory serves as the sole task-level motion command, while the controller resolves morphology-dependent base motion and retains arm-specific kinematic execution.
% A shared whole-body coordinator uses the robot state, arm morphology, and future EE poses to determine the required base motion.
% In parallel, an arm-specific IK module tracks the active EE target, while a frozen locomotion backend executes the base command.
A shared whole-body coordinator determines the required base motion, while an arm-specific IK module tracks the active EE target and a locomotion backend executes the resulting base command.
% The task policy therefore specifies only the desired EE trajectory, without providing base-motion commands.

Our key design insight is to use future task motion to anticipate base coordination, while combining static arm morphology with online interaction feedback to adapt that coordination across arms.
To extend reach, the coordinator uses trajectory preview to anticipate the need for base posture adjustment or locomotion.
% Training includes trajectories both within and beyond initial fixed-base reach.
% To support reuse across arms, we encode the known URDF arm model into a compact representation using an encoder pretrained with kinematic and dynamic supervision.
To support reuse across arms, we encode the known URDF arm model into a compact representation pretrained with kinematic and dynamic supervision, while retaining arm-specific IK for exact kinematic constraints.
% The same representation conditions the coordinator and locomotion backend through separate projections, while arm-specific IK handles joint limits and collision constraints.
% We further provide an arm-induced residual wrench to both learned modules and a short target--wrench--response history to the backend to account for arm--base interaction.
% This representation conditions the coordinator and locomotion backend through separate projections, while arm-specific IK handles joint limits and collision constraints.
% Online interaction feedback complements this static representation: the coordinator receives the current arm-induced residual wrench, while the backend receives a short target--wrench--response history.
Online residual-wrench feedback further captures realized arm--base interaction beyond this static model.

We evaluate \core in simulation and on a physical Go2--ARX5\footnote{ARX: \url{https://arx-x.com/}} system.
In simulation, we test held-out trajectories and mounted arms from four kinematic families using the same coordinator and frozen locomotion backend without policy fine-tuning.
% On hardware, \core connects to $\pi_0$~\cite{pi0} to perform language-conditioned manipulation beyond initial fixed-base reach.
On hardware, we demonstrate EE-commanded teleoperation and language-conditioned extended-reach pick-and-place with $\pi_0$~\cite{pi0}.
Our study uses a common Go2 base platform and considers known arm models and smooth, low-speed EE trajectories.

% The main contributions of this paper are as follows:
% \begin{itemize}
%     \item We present \core, a hierarchical WBC framework that converts short-horizon EE trajectories into coordinated arm and base motion without task-level base commands.

%     \item We use trajectory preview to coordinate base posture and locomotion for EE-trajectory tracking beyond initial fixed-base reach.

%     \item We combine URDF-derived morphology conditioning, arm-specific IK, and residual-wrench feedback to reuse the same learned controllers across held-out arms from four kinematic families.
% \end{itemize}

The main contributions of this paper are as follows:
\begin{itemize}
    \item We present \core, a shared whole-body control hierarchy that uses EE-trajectory preview to generate morphology-dependent arm--base motion within and beyond initial fixed-base reach, without task-level base commands.

    \item We combine URDF-derived morphology conditioning with online residual-wrench feedback to reuse the same learned coordinator and locomotion backend across mounted-arm morphologies, while retaining arm-specific IK.

    \item We evaluate cross-arm and extended-reach tracking on held-out trajectories and arms from four kinematic families in simulation, and demonstrate language-conditioned manipulation with $\pi_0$ on a physical Go2--ARX5 system.
\end{itemize}
\section{Related work}
\label{sec:relatedwork}

\subsection{Whole-Body Control}

% Traditional WBC coordinates multiple objectives through operational-space control and task hierarchies~\cite{operational_space,sentis_wbc}.
% For quadrupedal mobile manipulators, ALMA~\cite{alma} coordinates articulated locomotion and manipulation, while a unified model predictive control (MPC) framework~\cite{unified_mpc} jointly optimizes base, contact, and arm motion.
% Related optimization-based methods address motion and force planning, collision avoidance, and multicontact manipulation~\cite{stormmap,collision_free_wb_mpc,versatile_multicontact}.
% Reachability priors have also been used to guide mobile-base placement~\cite{reachability_priors}.

Traditional WBC coordinates multiple objectives through operational-space control and task hierarchies~\cite{operational_space,sentis_wbc}.
For quadrupedal mobile manipulators, ALMA~\cite{alma} coordinates articulated locomotion and manipulation, while unified model predictive control (MPC)~\cite{unified_mpc} jointly optimizes base, contact, and arm motion.
Optimization-based methods also address motion and force planning, collision avoidance, and multicontact manipulation~\cite{stormmap,collision_free_wb_mpc,versatile_multicontact}.
Reachability priors also guide mobile-base placement~\cite{reachability_priors}.

Learning-based methods instead acquire whole-body coordination from simulation.
DeepWBC~\cite{deep_wbc} learns a unified arm--leg policy, VBC~\cite{visual_wbc} tracks EE targets together with base velocities, and RoboDuet~\cite{roboduet} coordinates separate locomotion and arm policies.
UMI-on-Legs~\cite{umi_on_legs} uses task-frame EE trajectories, while MLM~\cite{mlm} extends this interface to multiple tasks.
Zhou et al.~\cite{tactile_aware_loco_manipulation} add tactile targets, PAKE~\cite{pake} generates partial kinematic references for low-level imitation, and CMP~\cite{cmp} addresses commands outside a fixed controller's nominal competence region.
These approaches primarily consider nearby targets and a specific type of robotic arm. In contrast, \core studies trajectory tracking within and beyond fixed-base reach across different arm morphologies.

\subsection{Cross-Morphology Control}

Robust locomotion across varying dynamics is commonly addressed through adaptation or morphology conditioning.
RMA~\cite{rma} infers latent environment properties from recent observations, while Ma et al.~\cite{hybrid_loco_manipulation} condition locomotion on predicted external wrenches and demonstrate transfer to unseen manipulators.
Manipulator-induced disturbances have also been addressed through disturbance predictive control~\cite{disturbance_predictive_loco_manipulation}.
GenLoco~\cite{genloco} trains shared locomotion policies over procedurally randomized quadruped morphologies, while LocoFormer~\cite{locoformer} combines large-scale procedural robot generation with long-context adaptation to generalize across embodiments.
Shared policies can exploit articulated graphs and repeated modules across different robot structures~\cite{nervenet,shared_modular_policies}. MetaMorph~\cite{metamorph} and ModuMorph~\cite{modumorph} use morphology information to support policy sharing.
% In \core, we fix a compact quadrupedal base with tight whole-body constraints and vary the mounted arm, combining static morphology information with arm-specific IK and online interaction feedback.
In \core, we instead fix a compact quadrupedal base with tight whole-body constraints and vary the mounted arm, combining explicit static morphology information with arm-specific IK and online interaction feedback.

\subsection{Visuomotor Policies}

% Recent visuomotor policies commonly predict short action sequences.
% ACT~\cite{act} generates action chunks with a Transformer, while Diffusion Policy~\cite{diffusion_policy} models action sequences for receding-horizon control.
% UMI~\cite{umi} adopts relative EE trajectories, and Mobile ALOHA~\cite{mobile_aloha} learns mobile manipulation from whole-body teleoperation.
% Generalist policies such as Octo~\cite{octo} and VLA policies such as RT-2~\cite{rt2}, OpenVLA~\cite{openvla}, and $\pi_0$~\cite{pi0} connect visual observations and task instructions to robot actions.
% CAC-VLA~\cite{cacvla} introduces coarse-to-fine latent-action conditioning for VLA action generation.
% DriftingVLA~\cite{driftingvla} enables one-step VLA action-chunk generation through per-dimension temporal drifting.
% FAVLA~\cite{favla} decouples slow VLM planning from fast force-aware action generation for contact-rich manipulation.
% STDArm~\cite{stdarm} transfers visuomotor policies trained on static data to moving platforms through real-time action correction.
% DreamTrajectory~\cite{dreamtrajectory} jointly predicts an EE trajectory and a whole-body action chunk.
% Rather than introducing another task policy, \core provides a downstream controller that converts EE trajectories into morphology-dependent arm and base motion.

Recent visuomotor policies commonly predict short action sequences.
ACT~\cite{act} generates action chunks with a Transformer, while Diffusion Policy~\cite{diffusion_policy} models action sequences for receding-horizon control.
Generalist policies such as Octo~\cite{octo} and VLA policies such as RT-2~\cite{rt2}, OpenVLA~\cite{openvla}, and $\pi_0$~\cite{pi0} further connect visual observations and task instructions to robot actions.
Recent variants improve action generation through coarse-to-fine latent conditioning~\cite{cacvla}, one-step temporal drifting~\cite{driftingvla}, or force-aware fast--slow control~\cite{favla}.
At the interface and execution level, UMI~\cite{umi} adopts relative EE trajectories, while Mobile ALOHA~\cite{mobile_aloha} learns mobile manipulation from whole-body teleoperation.
STDArm~\cite{stdarm} transfers visuomotor policies trained on static data to moving platforms through real-time action correction, and DreamTrajectory~\cite{dreamtrajectory} jointly predicts an EE trajectory and a whole-body action chunk.
These works primarily address task-level action generation, representation, or policy transfer.
In contrast, \core takes a short-horizon EE trajectory as the task-level command and focuses on its downstream realization as morphology-dependent arm and base motion.

\section{Methodology}
\label{sec:methodology}

\begin{figure*}[t]
    \centering
    \includegraphics[width=\textwidth]{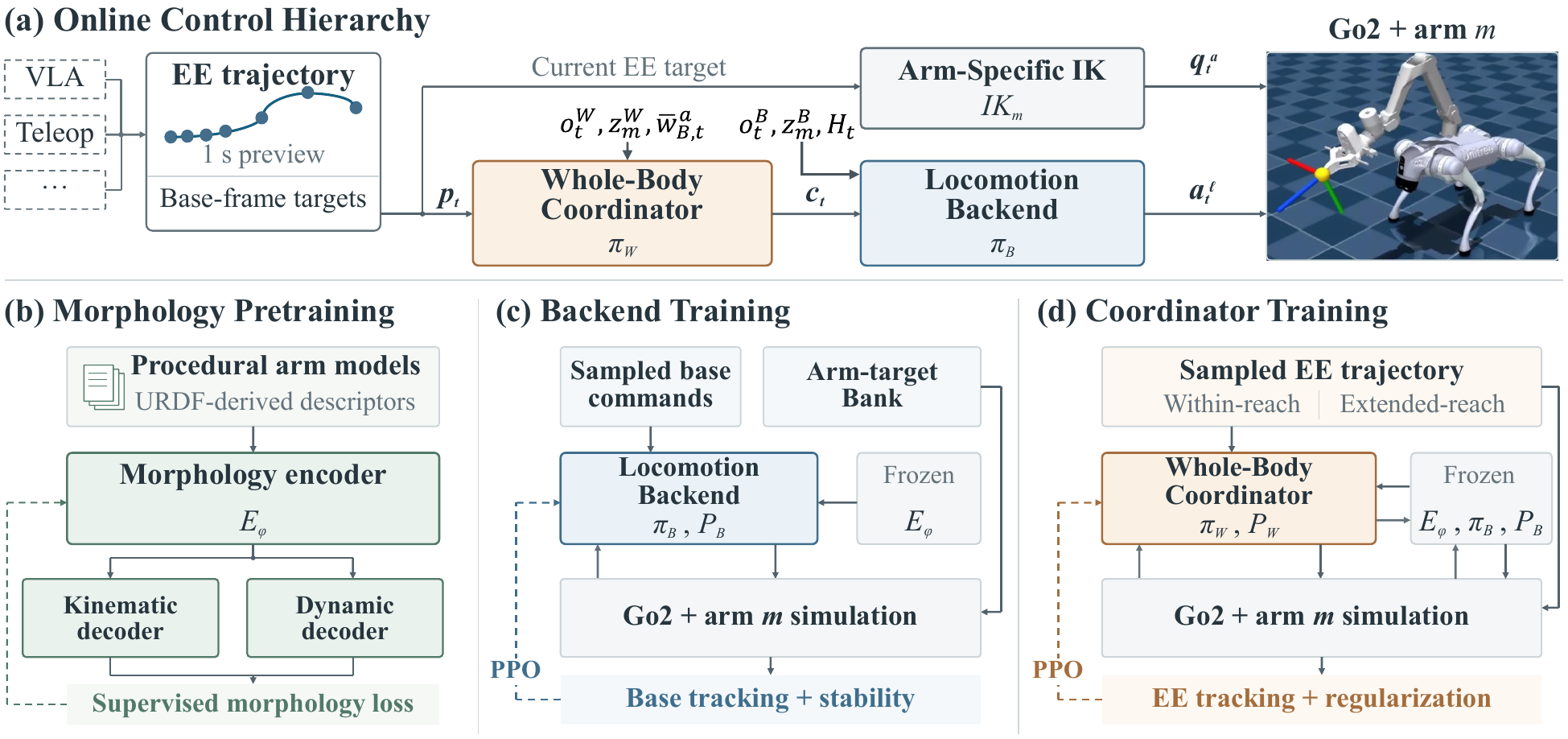}
    % \caption{\textbf{Pipeline of \core.}
    % (a) VLA policies and teleoperation are example sources of task-level EE trajectories, which are expressed in the current base frame. The whole-body coordinator uses the trajectory preview $\mathbf{p}_t$ to generate the base command $\mathbf{c}_t$, while arm-specific IK tracks the active EE pose. The locomotion backend maps $\mathbf{c}_t$ to leg actions $\mathbf{a}^{\ell}_t$.
    % (b) A shared morphology encoder maps the URDF-derived descriptor to separate conditioning vectors for the coordinator and backend.
    % (c) Residual force and torque form the current residual wrench $\bar{\mathbf{w}}^{a}_{B,t}$. Together with the active target and base response, it forms the target--wrench--response history $\mathbf{H}_t$ used by the backend, while the current residual wrench also conditions the coordinator.}
    \caption{\textbf{Pipeline of \core.}
    (a) The coordinator converts an EE trajectory preview into a base command, while arm-specific IK tracks the current target and the locomotion backend produces leg actions.
    (b) The morphology encoder is pretrained with kinematic and residual-wrench supervision and then frozen.
    (c) The backend and its morphology projection are trained with PPO under sampled base commands and independently executed, offline-IK-based arm motion, using target--wrench--response history.
    (d) With the encoder and backend frozen, the coordinator and its projection are trained through sampled EE trajectory covering motion within and beyond initial fixed-base reach.
    Dashed arrows indicate optimization updates.}
    \label{fig:pipeline}
    \vspace{-1.0em}
\end{figure*}

Given a short-horizon EE trajectory and the known model of the mounted arm, our goal is to generate coordinated arm and base motion without requiring task-level base-motion commands.
% To this end, \core combines a whole-body coordinator, an arm-specific IK solver, and a locomotion backend.
% We first introduce the complete hierarchy and then describe its morphology representation, locomotion backend, and whole-body coordinator.
% Fig.~\ref{fig:pipeline} summarizes the resulting hierarchy: the EE trajectory drives learned base coordination and arm-specific kinematic execution, while static morphology context and online interaction feedback condition the shared learned modules.
Fig.~\ref{fig:pipeline} summarizes the online control hierarchy in (a) and the training of the morphology encoder, locomotion backend, and whole-body coordinator in (b)--(d).

\subsection{Overview}
\label{sec:method_overview}

An EE trajectory specifies the desired EE motion but not its realization by the mounted arm and mobile base. 
% We therefore retain a common trajectory interface while separating learned base coordination from arm-specific kinematic execution. The quadruped platform and leg action space are shared across all instances, whereas $m$ indexes the mounted-arm morphology.
As shown in Fig.~\ref{fig:pipeline}(a), we retain a common trajectory interface while separating learned base coordination from arm-specific kinematic execution.
The quadruped platform and leg action space are shared across all instances, whereas $m$ indexes the mounted-arm morphology.

Let $B$ and $E$ denote the base, and EE frames, respectively, and let ${}^{B}\mathbf{T}_{E}\in SE(3)$ denote the pose of EE frame expressed in base frame.
At control step $t$, the input trajectory specifies $K$ desired EE poses at time offsets $0=\tau_0<\cdots<\tau_{K-1}$.
We express each pose in the current base frame and stack the resulting features:
\begin{equation}
\begin{aligned}
    \mathbf{g}_{t,k}
    &=\psi\!\left(
        % ({}^{R}\mathbf{T}_{B,t})^{-1}
        % {}^{R}\mathbf{T}_{E,t+\tau_k}
        {}^{B}\mathbf{T}_{E,t+\tau_k}
        % {}^{B}\mathbf{T}^{*}_{E,t+\tau_k}
    \right),\\
    \mathbf{p}_t
    &=\begin{bmatrix}
        \mathbf{g}_{t,0}^{\top}&\cdots&
        \mathbf{g}_{t,K-1}^{\top}
    \end{bmatrix}^{\top},
\end{aligned}
\label{eq:trajectory_preview}
\end{equation}
Here, $k\in\{0,\ldots,K-1\}$ indexes the preview poses and $\psi(\cdot)$ maps a relative EE pose to its translation and a continuous 6D rotation representation.
% where ${}^{B}\mathbf{T}^{*}_{E,t+\tau_k}$ denotes the \(k\)-th desired EE pose expressed in the current base frame and $\psi(\cdot)$ maps a relative EE pose to its translation and a continuous 6D rotation representation.
The resulting feature $\mathbf{g}_{t,k}\in\mathbb{R}^{9}$ describes one target, and $\mathbf{p}_t$ contains the current target at $\tau_0=0$ together with the future targets.

The complete hierarchy is
\begin{align}
    \mathbf{c}_t
    &=\pi_{\mathrm W}\!\left(
        \mathbf{o}^{\mathrm W}_t,
        \mathbf{p}_t,
        \bar{\mathbf{w}}^{a}_{B,t},
        \mathbf{z}^{\mathrm W}_m
    \right),
    \label{eq:coordinator_overview}\\
    \mathbf{q}^{a}_t
    &=\operatorname{IK}_{m}\!\left(
        % ({}^{R}\mathbf{T}_{B,t})^{-1}
        % {}^{R}\mathbf{T}_{E,t};
        {}^{B}\mathbf{T}_{E,t};
        \mathbf{q}^{a}_{\hat{t}}
    \right),
    \label{eq:ik_overview}\\
    \mathbf{a}^{\ell}_t
    &=\pi_{\mathrm B}\!\left(
        \mathbf{o}^{\mathrm B}_t,
        \mathbf{c}_t,
        \mathbf{H}_t,
        \mathbf{z}^{\mathrm B}_m
    \right),
    \label{eq:backend_overview}
\end{align}
where $\pi_{\mathrm W}$ is the whole-body coordinator, $\operatorname{IK}_{m}$ is the arm-specific IK solver, and $\pi_{\mathrm B}$ is the locomotion backend.
Superscripts $a$ and $\ell$ denote arm and leg quantities, and $\mathbf{q}^{a}_{\hat{t}}$ is the most recent arm-joint reference and is used to initialize the IK solver.
% The vectors $\mathbf{o}^{\mathrm W}_t$ and $\mathbf{o}^{\mathrm B}_t$ are the observations available to the two policies at deployment.
% The observation vectors \(\mathbf{o}^{\mathrm W}_t\) and \(\mathbf{o}^{\mathrm B}_t\) collect the inputs to the two policies and are detailed below.
\(\mathbf{o}^{\mathrm W}_t\) and \(\mathbf{o}^{\mathrm B}_t\) denote the observation vectors of the two policies and are detailed below.
The scaled residual wrench $\bar{\mathbf{w}}^{a}_{B,t}$ and target--wrench--response history $\mathbf{H}_t$ provide online interaction feedback, while $\mathbf{z}^{\mathrm W}_m$ and $\mathbf{z}^{\mathrm B}_m$ provide static morphology context.

The coordinator outputs the backend command:
\begin{equation}
    \mathbf{c}_t
    =
    \begin{bmatrix}
        v_{x,t} & v_{y,t} & \omega_{z,t} &
        \theta_{B,t} & h_{B,t}
    \end{bmatrix}^{\top},
    \label{eq:coordinator_command}
\end{equation}
where $v_{x,t}$ and $v_{y,t}$ are desired planar velocities, $\omega_{z,t}$ is the desired yaw rate, and $\theta_{B,t}$ and $h_{B,t}$ are the desired base pitch and height offsets relative to the nominal standing posture.
% The IK solver produces six arm-joint references $\mathbf{q}^{a}_t$, while the backend produces twelve leg-joint actions $\mathbf{a}^{\ell}_t$.
The IK solver produces the arm action $\mathbf{q}^{a}_t \in \mathbb{R}^{9}$, while the backend produces the leg action $\mathbf{a}^{\ell}_t \in \mathbb{R}^{12}$.
% Both branches use the same current EE target; gripper commands are passed separately and omitted here.
Gripper commands are passed separately and omitted here.

\subsection{Arm Morphology}
\label{sec:arm_morphology}

% A shared hierarchy must distinguish how the same EE motion affects different mounted arms. 
% A discrete arm identifier cannot represent an unseen morphology, whereas raw URDF fields mix geometric and inertial quantities with different relevance to the two control levels. 
% We therefore encode a canonical physical descriptor derived from the URDF.

% Arm morphology affects both reachable EE poses and disturbances transmitted to the mobile base.
Arm morphology affects both the kinematic feasibility of EE motion and the arm--induced disturbances.
% Fig.~\ref{fig:introduction}(a) illustrates the resulting differences in fixed-base workspace among representative arms.
% We therefore encode the known arm model into a shared representation for the coordinator and locomotion backend, as summarized in Fig.~\ref{fig:pipeline}(b).
We therefore learn a shared representation of the known arm model to condition the coordinator and locomotion backend. Fig.~\ref{fig:pipeline}(b) illustrates the encoder pretraining procedure.

\begin{table}[t]
    \centering
    % \caption{Kinematic families considered in this work.}
    \caption{Mounted-arm kinematic families and their anchor models.}
    \label{tab:arm_families}
    \small
    \setlength{\tabcolsep}{4pt}
    \begin{tabular}{@{}lll@{}}
        \toprule
        Code & Kinematic family & Anchor arm \\
        \midrule
        PA & Three parallel axes & ARX X5 \\
        SW & Spherical wrist     & PiPER \\
        OW & Offset wrist        & Compact xArm6 \\
        CW & Curved wrist        & Compact MICO \\
        \bottomrule
    \end{tabular}
    % \vspace{-1.0em}
\end{table}

% We consider the four 6-DoF kinematic families in Table~\ref{tab:arm_families}.
% Each family uses a distinct template for joint frames, joint axes, and wrist offsets, while procedural sampling varies link geometry, joint placement, and inertial properties.
% All arms follow the same mounting convention and canonical joint order, with a consistent definition of the tool center point (TCP).
% For morphology $m$, the URDF-derived descriptor $\boldsymbol{\theta}_m$ contains the mount and TCP transforms, ordered joint transforms, axes and limits, and link masses, centers of mass, and inertias.

% We consider four 6-DoF families, named by their dominant kinematic pattern and generated from anchor-specific templates of joint frames, axes, and wrist offsets (Table~\ref{tab:arm_families}). 
We consider four 6-DoF families chosen to span distinct kinematic layouts of mounted serial manipulators, whose different joint-axis arrangements and offsets induce substantially different workspaces, as shown in Table~\ref{tab:arm_families}.
Each family is defined by an anchor-specific template of joint frames, axes, and wrist offsets.
% Procedural sampling jointly varies joint placement, link geometry, and physically consistent inertial properties.
Within each family, procedural sampling jointly varies joint placement, link geometry, and physically consistent inertial properties, providing continuous morphology variation around a valid kinematic template.
% This exposes the shared controller to both cross-family changes in kinematic structure and within-family changes in geometry and dynamics, rather than only a small set of fixed arm instances.
% All morphologies share a fixed Go2 mount, canonical 6R joint order, TCP convention, and base-frame home pose.
All morphologies share a fixed Go2 mount, canonical 6R joint order, tool center point (TCP) convention, and base-frame home pose, keeping the control interface consistent while isolating variation in the mounted arm.
For morphology $m$, the URDF-derived descriptor $\boldsymbol{\theta}_m$ comprises the mount and TCP transforms; ordered joint transforms, axes, and limits; and per-link masses, centers of mass, and inertia tensors.

The shared encoder and the two projections are
\begin{equation}
\begin{aligned}
    \mathbf{e}_m
    &=E_{\phi}(\boldsymbol{\theta}_m)\in\mathbb{R}^{64},\\
    \mathbf{z}^{\mathrm B}_m
    &=P_{\mathrm B}(\hat{\mathbf{e}}_m)\in\mathbb{R}^{32},
    &
    \mathbf{z}^{\mathrm W}_m
    &=P_{\mathrm W}(\hat{\mathbf{e}}_m)\in\mathbb{R}^{32}.
\end{aligned}
\label{eq:morphology_embedding}
\end{equation}
Here, $E_{\phi}$ is the morphology encoder with parameters $\phi$, $\hat{\mathbf{e}}_m$ is its component-wise normalized output, and $P_{\mathrm B}$ and $P_{\mathrm W}$ are the backend and coordinator projections.
% (can be removed)
The shared embedding represents the arm model, while the projections adapt it to the two control levels.

To retain both kinematic and dynamic information, we pretrain $E_{\phi}$ with two supervised decoders:
\begin{equation}
    L_{\mathrm{morph}}
    =\sum_{s\in\{\mathrm{kin},\mathrm{dyn}\}}
    \lambda_s\,
    \mathbb{E}\!\left[
        \rho_{\delta}\!\left(
            D_s(\mathbf{e}_m,\mathbf{u}_s)-\mathbf{y}_s
        \right)
    \right].
    \label{eq:morphology_loss}
\end{equation}

For supervision type $s$, $D_s$, $\mathbf{u}_s$, and $\mathbf{y}_s$ denote the decoder, query, and target, respectively.
The expectation is over the pretraining queries, and $\rho_{\delta}$ is the mean component-wise Huber penalty with threshold $\delta$.
We set $\lambda_{\mathrm{kin}}=1$ and use $\lambda_{\mathrm{dyn}}$ to weight dynamic supervision.
Kinematic queries use normalized arm-joint configurations to predict the TCP position and two TCP-axis marker positions.
Dynamic queries contain the arm-joint configuration $\mathbf{q}$, velocity $\dot{\mathbf{q}}$, acceleration $\ddot{\mathbf{q}}$, and base-frame gravity $\mathbf{g}^{B}$; their target is the standardized residual wrench derived from Eq.~\eqref{eq:residual_wrench}.
After pretraining, the decoders are discarded and $E_{\phi}$ is frozen.
The embedding supplies morphology context but does not replace arm-specific IK or collision checking.

\subsection{Residual-Wrench-Aware Locomotion Backend}
\label{sec:locomotion_backend}

The URDF describes the nominal arm morphology, but the disturbance transmitted to the mobile base also depends on the instantaneous arm motion, tracking error, and model mismatch. We therefore complement the static morphology embedding with an online arm-induced residual wrench and its recent response history.

Let $O_B$ be a fixed reference point in frame $B$, and let $N_a$ be the number of arm links.
For link $i$, $m_i$ is its mass, $\mathbf{r}^{B}_{i,t}$ is its center-of-mass position relative to $O_B$, and $\mathbf{a}^{B}_{i,t}$ is the corresponding finite-difference acceleration relative to the base.
We compute the residual wrench as
\begin{equation}
\begin{aligned}
    \mathbf{w}^{a}_{B,t}
    &=\begin{bmatrix}
        \mathbf{f}^{a}_{B,t}\\
        \boldsymbol{\tau}^{a}_{B,t}
    \end{bmatrix},
    \quad
    \mathbf{f}^{a}_{B,t}
    =-\sum_{i=1}^{N_a}m_i\mathbf{a}^{B}_{i,t},\\
    \boldsymbol{\tau}^{a}_{B,t}
    &=\sum_{i=1}^{N_a}
      \mathbf{r}^{B}_{i,t}\!\times\!(-m_i\mathbf{a}^{B}_{i,t})
    +\left(
        \sum_{i=1}^{N_a}m_i\mathbf{r}^{B}_{i,t}
        -m_0\mathbf{c}^{B}_0
      \right)\!\times\!\mathbf{g}^{B}_t.
\end{aligned}
\label{eq:residual_wrench}
\end{equation}
Here, $\mathbf{f}^{a}_{B,t}$ and $\boldsymbol{\tau}^{a}_{B,t}$ are residual force and torque, and $m_0$ and $\mathbf{c}^{B}_0$ are the nominal payload mass and center of mass relative to $O_B$.
All arms use the same frame, reference point, sign convention, and force--torque ordering.
The policy input is
\begin{equation}
    \bar{\mathbf{w}}^{a}_{B,t}
    =
    \begin{bmatrix}
        \mathbf{f}^{a}_{B,t}/s_f\\
        \boldsymbol{\tau}^{a}_{B,t}/s_{\tau}
    \end{bmatrix},
    \label{eq:wrench_normalization}
\end{equation}
where $s_f$ and $s_{\tau}$ are fixed numerical scales rather than physical bounds; the wrench is not hard clipped.

The current wrench alone does not describe how the mobile base has responded to recent arm motion.
We therefore retain a short target--wrench--response history:
\begin{align}
    \mathbf{h}_t
    &=\begin{bmatrix}
        \mathbf{g}_{t,0}^{\top}&
        (\bar{\mathbf{w}}^{a}_{B,t})^{\top}&
        (\boldsymbol{\omega}^{B}_t)^{\top}&
        (\hat{\mathbf{g}}^{B}_t)^{\top}
    \end{bmatrix}^{\top},\\
    \mathbf{H}_t
    &=\begin{bmatrix}
        \mathbf{h}_{t-N_h+1}^{\top}&\cdots&\mathbf{h}_{t}^{\top}
    \end{bmatrix}^{\top},
    \label{eq:response_history}
\end{align}
where $\boldsymbol{\omega}^{B}_t$ is the base angular velocity, $\hat{\mathbf{g}}^{B}_t$ is the projected gravity direction, and $N_h$ is the history length.
% Thus, $\mathbf h_t$ collects the current target, current residual wrench, and instantaneous base response, while $\mathbf H_t$ stacks these features over the most recent $N_h$ control steps.
The remaining backend observation $\mathbf{o}^{\mathrm B}_t$ contains the current leg-joint positions and velocities, the previous leg action, and the gait phase.
% Together with the static morphology context $\mathbf{z}^{\mathrm B}_m$, this history allows the backend to account for morphology- and motion-dependent arm disturbances while preserving a common locomotion-command interface. 
% We train $P_{\mathrm B}$ and $\pi_{\mathrm B}$ across the procedural arms, and freeze both before training the whole-body coordinator.

As illustrated in Fig.~\ref{fig:pipeline}(c), we train $P_B$ and $\pi_B$ from scratch with proximal policy optimization (PPO)~\cite{ppo}, keeping $E_\phi$ frozen.
The backend tracks sampled base-velocity and posture commands while each arm independently executes minimum-jerk joint trajectories generated from a morphology-specific bank of offline IK solutions.
The target configurations are screened for joint margins and collisions.
Training randomizes base dynamics, friction, and actuator gains, and the objective emphasizes base-command tracking and stability with leg-motion regularization.

\subsection{Whole-Body Coordinator}
\label{sec:whole_body_coordinator}

% The locomotion backend executes a specified base command, but the current EE target alone does not indicate whether subsequent motion remains inside the arm workspace. 
% We therefore provide the coordinator with the preview in Eq.~\eqref{eq:trajectory_preview} so that it can anticipate the need for posture adjustment or locomotion.

While the locomotion backend tracks a given base command, the whole-body coordinator determines that command from the EE trajectory.
Since the current target alone does not indicate whether upcoming targets will remain within the arm's workspace, the coordinator uses the preview in Eq.~\eqref{eq:trajectory_preview} to anticipate the need for posture adjustment or locomotion.

% The coordinator is defined by Eq.~\eqref{eq:coordinator_overview}.
% Its observation $\mathbf{o}^{\mathrm W}_t$ contains the base angular velocity and projected gravity, the current EE tracking error, the current backend command, the previous coordinator action, and the morphology-normalized arm-joint configuration.
% The preview $\mathbf{p}_t$ provides future task information, while $\bar{\mathbf{w}}^{a}_{B,t}$ and $\mathbf{z}^{\mathrm W}_m$ provide the current arm-induced disturbance and static morphology context, respectively.
% In parallel, the current target is passed to the arm-specific solver in Eq.~\eqref{eq:ik_overview}, which uses the kinematics and joint limits of morphology $m$.
% This separation leaves arm-specific kinematic execution to IK while the shared coordinator determines the accompanying base motion.

The coordinator is defined by Eq.~\eqref{eq:coordinator_overview}.
The observation $\mathbf{o}_t^W$ contains the base angular velocity and projected gravity, the current EE tracking error, the current backend command, the previous coordinator action, and the morphology-normalized arm-joint configuration.

We train $P_W$ and $\pi_W$ from scratch with PPO while keeping $E_\phi$, $P_B$, and $\pi_B$ frozen, as illustrated in Fig.~\ref{fig:pipeline}(d).
The objective combines coarse and fine EE pose tracking, IK command quality, base--EE heading alignment, base stability, economical base motion, and joint-, contact-, and collision-related penalties.

% Training samples EE trajectory from a common home pose, covering continuations, turns, and reversals within and beyond initial fixed-base reach.
% Training EE trajectories are sampled from a fixed procedural bank of three-pose paths starting from a common home pose and covering continuations, turns, and reversals within and beyond the initial fixed-base reach.
% Per-episode stratified sampling balances ordinary tracking and extended-reach paths, whose sampling weight increases during training.
Training EE trajectories are sampled from a fixed procedural bank of three-pose continuations, turns, and reversals starting from a common home pose, and are divided into within-reach and extended-reach subsets according to whether they remain within the initial fixed-base reach.
Per-episode stratified sampling draws from both subsets, with the probability of extended-reach trajectories increased over training.
Positions use piecewise cubic interpolation, while orientations use quaternion SLERP through the three knots.
Offline IK evaluations define family-specific height--pitch envelopes based on feasibility at admissible horizontal placements; broad planar target ranges leave base repositioning to the coordinator.
% Held-out completion statistics inform narrower family-specific deployment envelopes, providing empirical operating limits rather than guarantees of whole-body feasibility.
Held-out completion statistics inform narrower family-specific envelopes only for real-world deployment, providing empirical operating limits rather than guarantees of whole-body feasibility.

\section{Experiments}
\label{sec:experiment}

\begin{table*}[t]
    \centering
    % \caption{Evaluation on held-out trajectories and morphologies.}
    \caption{
        EE-trajectory tracking on held-out trajectories and arm morphologies. The two subsets contain trajectories with at least one reference target satisfying the indicated $x$ condition in the initial base frame.
    }
    \label{tab:simulation_main}
    \small
    \setlength{\tabcolsep}{5pt}
    \begin{tabular}{@{}lccccccccc@{}}
        \toprule
        & \multicolumn{3}{c}{All trajectories}
        & \multicolumn{3}{c}{Extended reach ($x>1\,\mathrm{m}$)}
        & \multicolumn{3}{c}{Extended reach ($x<0\,\mathrm{m}$)} \\
        \cmidrule(lr){2-4}
        \cmidrule(lr){5-7}
        \cmidrule(lr){8-10}
        Family
        & $R_{\mathrm{comp}}$ (\%) $\uparrow$
        & $e_p$ (cm) $\downarrow$
        & $e_R$ ($^\circ$) $\downarrow$
        & $R_{\mathrm{comp}}$ (\%) $\uparrow$
        & $e_p$ (cm) $\downarrow$
        & $e_R$ ($^\circ$) $\downarrow$
        & $R_{\mathrm{comp}}$ (\%) $\uparrow$
        & $e_p$ (cm) $\downarrow$
        & $e_R$ ($^\circ$) $\downarrow$ \\
        \midrule
        PA      & 98.45 & 1.62 & 3.06 & 97.57 & 1.35 & 2.43 & 99.04 & 1.80 & 3.12 \\
        SW      & 98.65 & 1.96 & 4.10 & 99.65 & 1.58 & 3.05 & 97.81 & 2.03 & 4.09 \\
        OW      & 91.03 & 2.75 & 6.40 & 94.43 & 2.12 & 4.81 & 88.39 & 2.60 & 5.29 \\
        CW      & 87.65 & 2.88 & 5.68 & 87.65 & 2.70 & 5.15 & 81.01 & 3.20 & 6.23 \\
        \midrule
        Overall & 93.95 & 2.30 & 4.81 & 94.83 & 1.94 & 3.86 & 91.56 & 2.41 & 4.68 \\
        \bottomrule
    \end{tabular}
    % \vspace{-1.0em}
\end{table*}
\begin{table}[t]
    \centering
    \caption{EE-trajectory tracking on anchor arm models without policy fine-tuning.}
    \label{tab:anchor_arms}
    \small
    \setlength{\tabcolsep}{3pt}
    \begin{tabular}{@{}lccc@{}}
        \toprule
        Anchor arm (family)
        & $R_{\mathrm{comp}}$ (\%) $\uparrow$
        & $e_p$ (cm) $\downarrow$
        & $e_R$ ($^\circ$) $\downarrow$ \\
        \midrule
        ARX X5 (PA)        & 99.30 & 1.57 & 2.96 \\
        PiPER (SW)         & 93.16 & 1.98 & 4.45 \\
        Compact xArm6 (OW) & 99.27 & 1.93 & 4.09 \\
        Compact MICO (CW)  & 86.62 & 2.67 & 5.26 \\
        \bottomrule
    \end{tabular}
    % \vspace{-1.0em}
\end{table}
\begin{table}[t]
    \centering
    \caption{Comparison of EE-trajectory tracking errors on Go2--ARX5 in MuJoCo.}
    \label{tab:mujoco_tracking}
    \begin{tabular}{lcccc}
        \toprule
        & \multicolumn{2}{c}{\core}
        & \multicolumn{2}{c}{UMI-on-Legs~\cite{umi_on_legs}} \\
        \cmidrule(lr){2-3}
        \cmidrule(lr){4-5}
        Trajectory shape
        & $e_p$ (cm) $\downarrow$ & $e_R$ ($^\circ$) $\downarrow$
        & $e_p$ (cm) $\downarrow$ & $e_R$ ($^\circ$) $\downarrow$ \\
        \midrule
        Vertical circle
            & 1.38 & 2.06
            & 5.00 & 22.72 \\
        Vertical square
            & 1.43 & 2.14
            & 5.17 & 23.76 \\
        Horizontal rectangle
            & 1.48 & 2.01
            & 4.77 & 40.41 \\
        Horizontal base orbit
            & 2.04 & 2.38
            & 7.31 & 76.57 \\
        \bottomrule
    \end{tabular}
    % \vspace{-1.0em}
\end{table}

% We evaluate \core with four questions in mind: whether it can track held-out EE trajectories, whether the same learned hierarchy can be reused across mounted-arm morphologies, whether morphology conditioning contributes to this generalization, and whether the EE-trajectory interface can connect a vision-language-action policy to a physical quadrupedal mobile manipulator.

We evaluate extended reach, transfer to unseen trajectories and mounted arms, the contribution of morphology conditioning, and the EE-trajectory interface on hardware.

\subsection{Experimental Setup}
\label{sec:experimental_setup}

% We fix the mobile base to a Unitree Go2 and procedurally generate 6-DoF mounted arms from the four kinematic families in Table~\ref{tab:arm_families}.
% The dataset contains 1024 arms for training, 128 for validation, and 128 for evaluation, with no overlap between the splits.
% The reference models PiPER, ARX X5, xArm6, and MICO are excluded from all stages of controller training and used only for held-out evaluation.

% \paragraph{Morphologies and trajectories}
% We use a Unitree Go2 as the common mobile base and procedurally generate mounted 6-DoF arms from the four kinematic families in Table~\ref{tab:arm_families}.
% For each family, we generate 256 training, 32 validation, and 32 held-out morphologies, yielding 1024, 128, and 128 morphology-disjoint instances, respectively.
% The exact anchor models are excluded from these splits and from policy optimization. Their evaluation measures transfer to anchor morphologies.
% The coordinator uses disjoint banks of 100k training and 10k held-out trajectories, constructed as described in Sec.~\ref{sec:whole_body_coordinator}. The trajectories spans $x\in[-0.32,1.21]$,m, $y\in[-0.66,0.66]$,m, and $z\in[-0.25,0.50]$,m, expressed in the initial base frame.

\paragraph{Morphologies and trajectories}
We use a Unitree Go2 as the common mobile base and procedurally generate mounted 6-DoF arms from the four kinematic families in Table~\ref{tab:arm_families}.
% Within each family, we vary link geometry and inertial properties around a physical template while preserving its joint-axis topology.
For each family, we generate 256 training, 32 validation, and 32 held-out morphologies, yielding 1024, 128, and 128 morphology-disjoint instances, respectively.
The exact anchor models are excluded from the generated splits and policy optimization, and are evaluated separately to assess transfer from their corresponding procedural families.
The coordinator uses disjoint banks of 100k training and 10k held-out trajectories, constructed as described in Sec.~\ref{sec:whole_body_coordinator}.
Across the four families, the trajectory envelope spans $x\in[-0.35,1.25]\,\mathrm{m}$, $y\in[-0.66,0.66]\,\mathrm{m}$, and $z\in[-0.25,0.50]\,\mathrm{m}$ in the initial base frame.
Arm commands are realized by morphology-specific bounded DLS IK with the corresponding joint limits.

\paragraph{Training and simulation protocol}
All policies are trained in IsaacLab~\cite{isaaclab}, following the staged procedure described in Sec.~\ref{sec:methodology}.
The coordinator uses $K=7$ preview poses spanning 1\,s, and the backend uses $N_h=5$ history frames.
The coordinator runs at 10\,Hz, while the backend and arm IK run at 50\,Hz.
The backend and coordinator are trained with 2,048 parallel environments for 12,000 and 2,000 PPO iterations, respectively.
% To verify that the learned behavior does not depend on the training simulator alone, we additionally evaluate the same policies in MuJoCo~\cite{mujoco} without retraining.
% To assess cross-simulator transfer, we additionally evaluate the same policies in MuJoCo~\cite{mujoco} without retraining.
% To verify that the learned behavior does not depend on the training simulator alone,
For cross-simulator evaluation,
we evaluate \core in MuJoCo~\cite{mujoco} without retraining and compare it with UMI-on-Legs~\cite{umi_on_legs} under the same task settings.
For UMI-on-Legs, we use the official implementation and the released random-trajectory checkpoint, without task-specific fine-tuning.

\paragraph{Real-world deployment}
For real-world evaluation, we deploy the same control hierarchy on a Go2 equipped with an ARX X5 arm.
EE motion commands are supplied through teleoperation or by $\pi_0$~\cite{pi0}, without task-level base-motion commands.
% Teleoperation inputs from [device] are converted into the EE trajectory preview using [implemented method].
% For the language-conditioned extended-reach pick-and-place task, $\pi_0$ provides short-horizon EE trajectories and gripper commands.
We collect 100 demonstrations for this task using a Pika handheld device and fine-tune $\pi_0$ following the UMI-style pipeline~\cite{umi}, with a wrist-mounted fisheye camera as the visual input for a wider field of view.
At deployment, $\pi_0$ runs on an external RTX 5090 workstation and communicates with the robot over wired Ethernet.

\paragraph{Evaluation metrics}
For simulation evaluation, we report three metrics: trajectory completion rate $R_{\mathrm{comp}}$, mean EE position error $e_p$, and mean EE orientation error $e_R$. 
$R_{\mathrm{comp}}$ is the fraction of evaluation episodes that finish the prescribed trajectory without tracking-failure or safety termination. 
Tracking failure is triggered by EE position error exceeding 0.20\,m after a 0.5\,s initialization grace period; safety terminations include unintended contact and loss of base stability. 
For each episode, we compute
\begin{equation}
    \begin{aligned}
        e_p &=
        \frac{1}{T}\sum_{t=1}^{T}
        % \left\|\mathbf p^d_{E,t}-\mathbf p_{E,t}\right\|_2,\\
        \left\| 
        \mathbf r^{\mathrm{des}}_{t} 
        - 
        \mathbf r^{\mathrm{act}}_{t}
        \right\|_2,\\
        e_R &=
        \frac{2}{T}\sum_{t=1}^{T}
        \arccos\!\left(
        \left|\left\langle \mathbf q^{\mathrm{des}}_{t},\mathbf q^{\mathrm{act}}_{t}\right\rangle\right|
        \right).
    \end{aligned}
    \label{eq:ee_tracking_metric}
\end{equation}
% Here $q^d_{E,t}$ and $q_{E,t}$ are unit orientation quaternions, and $T$ is the number of samples recorded before completion or termination.
Here, $\mathbf r^{\mathrm{des}}_t$ and $\mathbf r^{\mathrm{act}}_t$ denote the desired and actual EE positions, respectively, while $\mathbf q^{\mathrm{des}}_t$ and $\mathbf q^{\mathrm{act}}_t$ are the corresponding unit orientation quaternions. %$T$ is the number of samples recorded before completion or termination.
% We average each error over all evaluation episodes, including failures, and report position and geodesic orientation errors in cm and degrees, respectively.
We average these per-episode errors over all evaluation episodes, including failures.

\begin{figure}[t]
    \centering
    \includegraphics[width=\linewidth]{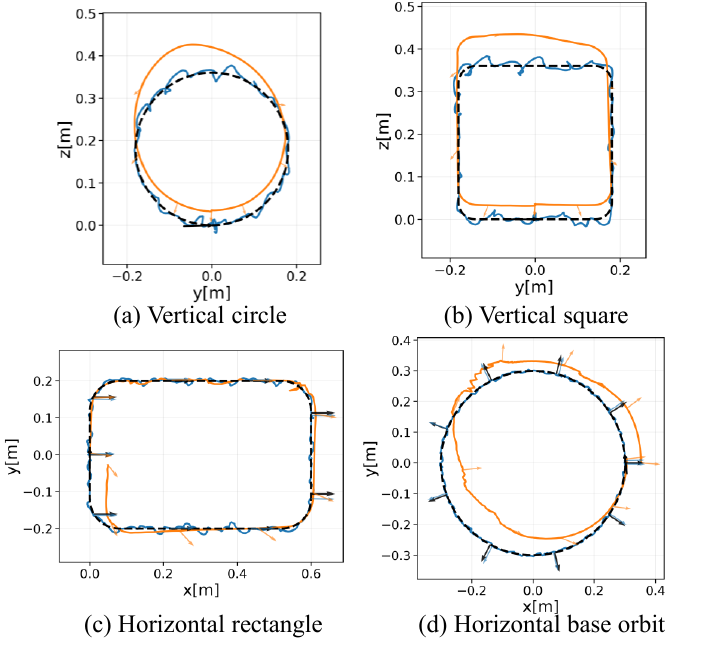}
    \caption{
        EE-trajectory tracking on Go2--ARX5 in MuJoCo.
        Panels (a)--(d) show the vertical circle, vertical square, horizontal rectangle, and horizontal base orbit, respectively.
        Black dashed curves show commanded trajectories; blue and orange solid curves show trajectories executed by \core{} and UMI-on-Legs, respectively.
        Color-matched arrows show EE local $z$-axes projected onto each plotting plane.
    }
    \label{fig:mujoco_tracking}
    % \vspace{-2.0em}
\end{figure}

\subsection{Simulation Evaluation}
\label{sec:simulation_evaluation}

\paragraph{Trajectory and morphology generalization}
We first evaluate \core on held-out trajectories and mounted-arm morphologies excluded from training. 
% Table~\ref{tab:simulation_main} reports the EE tracking error and trajectory completion rate for each kinematic family. 
% To assess reach extension through base motion, we additionally consider a extended-reach subset containing trajectories whose reference targets reach more than 1.0\,m ahead of the initial base position.
% To assess reach extension through base motion, we additionally consider a conservative extended-reach subset. We select trajectories containing a reference target more than 1.0\,m ahead of the initial base position; under the fixed-base configuration, such targets are infeasible for all arm morphologies considered in this study. This provides a morphology-independent sufficient condition for extended reach and simplifies subset construction.
To assess reach extension through base motion, we additionally consider two extended-reach subsets. We select trajectories containing a reference target more than $1.0\,\mathrm{m}$ ahead of the initial base position ($x>1\,\mathrm{m}$), or behind it ($x<0$). These provide morphology-independent geometric criteria and simplify subset construction.

Table~\ref{tab:simulation_main} reports tracking accuracy and completion rates for both the full evaluation set and these subsets.
% Across the four families, \core achieves a mean position error of 2.91\,cm and a trajectory completion rate of 85.65\%.
% Performance remains family-dependent, with completion ranging from 71.92\% for curved-wrist arms to 96.64\% for offset-wrist arms. On the extend-reach subset, \core achieves a completion rate of 81.57\% and a mean position error of 2.77\,cm.
% Completion remains comparable to the full-set result for PA and SW, but decreases for OW and CW, indicating that performance remains family-dependent.
Across the four families, \core achieves a mean position error of $2.30\,\mathrm{cm}$, a mean orientation error of $4.81^\circ$, and a trajectory completion rate of $93.95\%$.
On the extended-reach ($x>1\,\mathrm{m}$) subset, it achieves $94.83\%$ completion with mean position and orientation errors of $1.94\,\mathrm{cm}$ and $3.86^\circ$, respectively.
On the extended-reach ($x<0$) subset, it achieves $91.56\%$ completion with mean position and orientation errors of $2.41\,\mathrm{cm}$ and $4.68^\circ$, respectively.
Together, these results support a shared whole-body tracking capability that extends operational reach across the considered kinematic families.

\begin{figure*}[t]
    \centering
    \includegraphics[width=\textwidth]{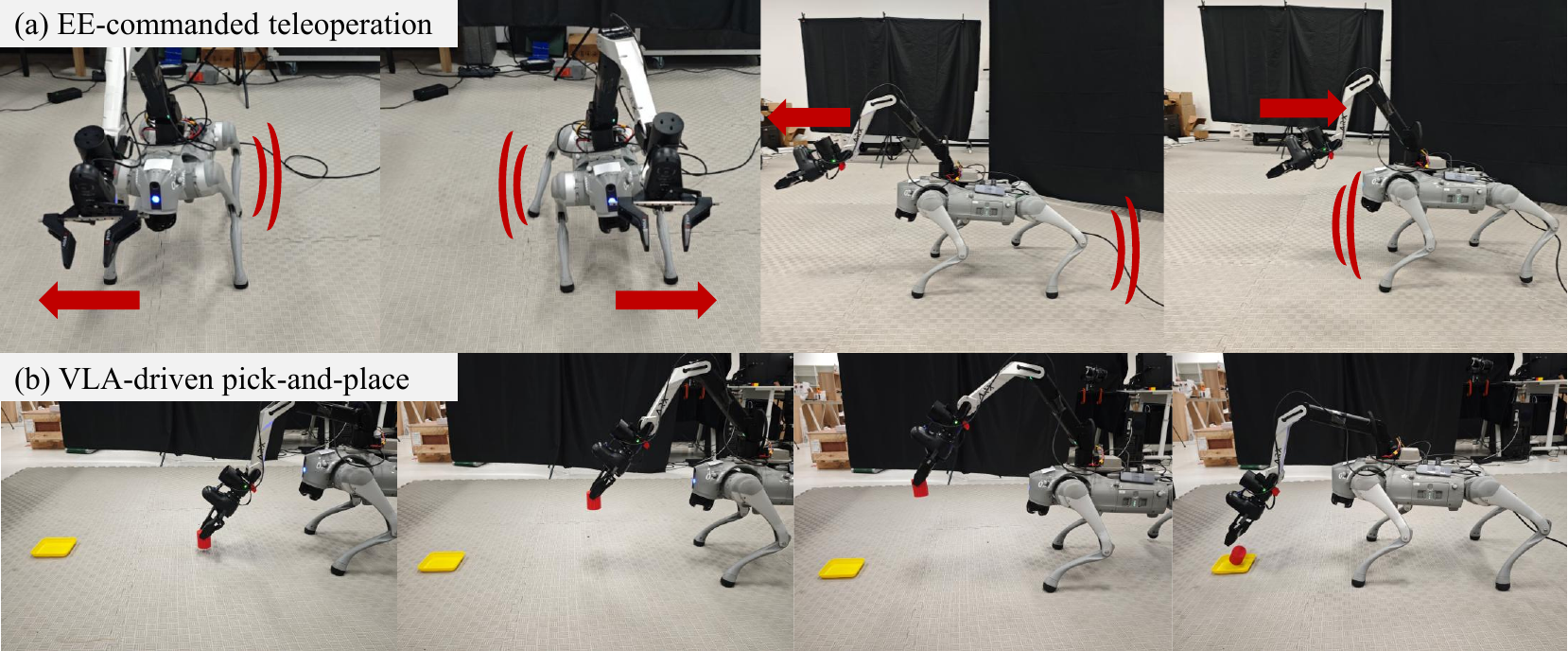}
    \caption{Real-world demonstrations of \core on a Go2--ARX5 platform.
    % (a) EE-commanded teleoperation with controller-generated base motion.
    (a) EE-commanded teleoperation, where straight red arrows indicate the commanded EE motion and curved red marks indicate the controller-generated base motion.
    (b) VLA-driven extended-reach pick-and-place using $\pi_0$.
    Each row shows successive stages from left to right.
    In both cases, only the EE motion is commanded at the task level; no base-motion command is provided by either the operator or $\pi_0$.}
    \label{fig:real_world}
    \vspace{-1.0em}
\end{figure*}

\paragraph{Transfer to anchor arms}
We further evaluate the anchor arm for each family, excluding the exact models from policy training.
Table~\ref{tab:anchor_arms} reports their position and orientation tracking errors and trajectory completion rates using the same learned hierarchy without policy adaptation.
Completion rates reach $99.30\%$ for ARX X5 and $99.27\%$ for Compact xArm6, with mean position errors of $1.57\,\mathrm{cm}$ and $1.93\,\mathrm{cm}$, respectively.
PiPER and Compact MICO achieve completion rates of $93.16\%$ and $86.62\%$, respectively.
% This evaluation complements the procedural held-out split by testing models corresponding to existing arm designs.
% Overall, the learned hierarchy generalizes well to the held-out anchor arms.
Overall, These results demonstrate the learned hierarchy generalizes well to the held-out anchor arms.

\paragraph{Cross-simulator validation and comparison}
We also evaluate \core in MuJoCo on four EE trajectories and compare it with UMI-on-Legs.
As shown in Table~\ref{tab:mujoco_tracking}, \core consistently achieves lower position and orientation errors, with mean position and orientation errors of $1.38$--$2.04$~cm and $2.01^\circ$--$2.38^\circ$, respectively, compared with $4.77$--$7.31$~cm and $22.72^\circ$--$76.57^\circ$ for UMI-on-Legs.
For the first two trajectories, which remain largely within the initial fixed-base reach, UMI-on-Legs keeps the base nearly stationary and tracks the positional path relatively smoothly, but with substantial tracking error.
This tendency is also reflected in the UMI-on-Legs ablation, where replacing task-specific UMI trajectories with random trajectories increases the mean orientation error from $3.35^\circ$ to $15.67^\circ$ on the authors' tossing benchmark~\cite{umi_on_legs}.
On the extended-reach trajectories, however, this tendency becomes more pronounced: UMI-on-Legs largely sacrifices orientation tracking and moves the base only reactively as the EE approaches its reach limit, producing intermittent corrections in the executed trajectory.
In contrast, \core actively coordinates base motion to maintain both position and orientation tracking, resulting in small local fluctuations while remaining close to the commanded trajectory.
Fig.~\ref{fig:mujoco_tracking} visualizes the resulting position and orientation tracking patterns through the executed EE trajectories and projected local $z$-axes.

\begin{table}[t]
    \centering
    \caption{Ablation on morphology conditioning on held-out arms. $\Delta R_{\mathrm{comp}}$ (pp) denotes the completion rate of the full model minus that of the variant without morphology conditioning.}
    \label{tab:ablation_study}
    \small
    \setlength{\tabcolsep}{5pt}
    \begin{tabular}{@{}lccc@{}}
        \toprule
        & \multicolumn{3}{c}{$\Delta R_{\mathrm{comp}}$ (pp)} \\
        \cmidrule(lr){2-4}
        Family
        & All trajectories
        & $x>1\,\mathrm{m}$
        & $x<0\,\mathrm{m}$ \\
        \midrule
        PA & $-0.48$ & $-1.22$ & $-0.14$ \\
        SW & $+0.55$ & $+0.52$ & $+0.41$ \\
        OW & $+2.59$ & $+2.26$ & $+3.18$ \\
        CW & $+9.90$ & $+9.52$ & $+8.24$ \\
        \bottomrule
    \end{tabular}
    \vspace{-1.0em}
\end{table}

\subsection{Ablation Study}
\label{sec:ablation}

We ablate morphology conditioning by training a variant without the morphology latent, keeping all other policy inputs, training trajectories, and optimization settings unchanged.
Table~\ref{tab:ablation_study} reports completion-rate differences on the same held-out arms and evaluation trajectories used above.
Morphology conditioning improves completion over all trajectories by 9.90 and 2.59 percentage points CW and OW, respectively, with positive gains in both position-based subsets.
% In contrast, the overall changes for PA and SW are below one percentage point in magnitude, suggesting that the benefit of morphology conditioning varies across arm families.
In contrast, the overall changes for PA and SW are below one percentage point in magnitude. 
% Overall, morphology conditioning improves the shared hierarchy beyond morphology-agnostic multi-arm training, enabling more effective adaptation across distinct arm families, particularly for OW and CW.
Overall, morphology conditioning improves the shared hierarchy beyond morphology-agnostic multi-arm training, enabling more effective adaptation across distinct arm families. 
% (can be removed)
The larger gains on the geometrically more distinct OW and CW families further suggest that explicit morphology information is particularly useful when a shared policy spans diverse kinematic structures and workspaces.

\subsection{Real-World Evaluation}
\label{sec:real_world}

Finally, we evaluate the EE-trajectory interface on hardware through teleoperation and a language-conditioned extended-reach pick-and-place task.
During teleoperation, the operator specifies EE motion, while \core generates the corresponding base motion and posture adjustments.
For pick-and-place, $\pi_0$ supplies EE trajectories and gripper commands, and \core coordinates arm and base motion to reach a required manipulation pose beyond the arm's initial fixed-base reach.
Neither the operator nor $\pi_0$ provides base-motion commands.
Fig.~\ref{fig:real_world} shows representative executions using the same control hierarchy.

\section{Conclusion}
\label{sec:conclusion}

We presented \core, a hierarchical framework that converts short-horizon EE trajectories into coordinated arm and base motion for quadrupedal mobile manipulation.
Trajectory preview enables manipulation beyond the initial fixed-base reach, while morphology conditioning, arm-specific IK, and residual-wrench feedback support reuse across mounted arms.
% Experiments on held-out trajectories and morphologies from four kinematic families, together with language-conditioned manipulation using $\pi_0$~\cite{pi0} on a physical Go2--ARX5 system, demonstrate the resulting generalization and applicability.
Simulation experiments demonstrate generalization to held-out trajectories and arm morphologies from four kinematic families. On a physical Go2--ARX5 system, EE-commanded teleoperation and language-conditioned extended-reach pick-and-place with $\pi_0$ show the applicability of the same EE-trajectory interface.
The current framework assumes a known arm model and smooth, low-speed EE trajectories on a common Go2 base; future work will extend it to different mobile bases and contact-rich manipulation.

\bibliographystyle{IEEEtran}
\bibliography{IEEEabrv, reference}

@inproceedings{alma,
  title={Alma-articulated locomotion and manipulation for a torque-controllable robot},
  author={Bellicoso, C Dario and Kr{\"a}mer, Koen and St{\"a}uble, Markus and Sako, Dhionis and Jenelten, Fabian and Bjelonic, Marko and Hutter, Marco},
  booktitle={International conference on robotics and automation},
  pages={8477--8483},
  year={2019},
  organization={IEEE}
}

@article{unified_mpc,
  title={A unified mpc framework for whole-body dynamic locomotion and manipulation},
  author={Sleiman, Jean-Pierre and Farshidian, Farbod and Minniti, Maria Vittoria and Hutter, Marco},
  journal={IEEE Robotics and Automation Letters},
  volume={6},
  number={3},
  pages={4688--4695},
  year={2021},
  publisher={IEEE}
}

@article{versatile_multicontact,
  title={Versatile multicontact planning and control for legged loco-manipulation},
  author={Sleiman, Jean-Pierre and Farshidian, Farbod and Hutter, Marco},
  journal={Science Robotics},
  volume={8},
  number={81},
  pages={eadg5014},
  year={2023},
  publisher={American Association for the Advancement of Science}
}

@inproceedings{deep_wbc,
  title={Deep whole-body control: Learning a unified policy for manipulation and locomotion},
  author={Fu, Zipeng and Cheng, Xuxin and Pathak, Deepak},
  booktitle={Conference on Robot Learning},
  pages={138--149},
  year={2023},
  organization={PMLR}
}

@inproceedings{visual_wbc,
  title={Visual whole-body control for legged loco-manipulation},
  author={Liu, Minghuan and Chen, Zixuan and Cheng, Xuxin and Ji, Yandong and Qiu, Ri-Zhao and Yang, Ruihan and Wang, Xiaolong},
  booktitle = {Conference on Robot Learning},
  pages     = {234--257},
  volume    = {270},
  year={2024},
  organization={PMLR}
}

@article{roboduet,
  title={Roboduet: Learning a cooperative policy for whole-body legged loco-manipulation},
  author={Pan, Guoping and Ben, Qingwei and Yuan, Zhecheng and Jiang, Guangqi and Ji, Yandong and Li, Shoujie and Pang, Jiangmiao and Liu, Houde and Xu, Huazhe},
  journal={IEEE Robotics and Automation Letters},
  volume={10},
  number={5},
  pages={4564--4571},
  year={2025},
  publisher={IEEE}
}

@inproceedings{umi_on_legs,
  title={Umi on legs: Making manipulation policies mobile with manipulation-centric whole-body controllers},
  author={Ha, Huy and Gao, Yihuai and Fu, Zipeng and Tan, Jie and Song, Shuran},
  booktitle = {Conference on Robot Learning},
  pages     = {5254--5270},
  volume    = {270},
  year={2024},
  organization={PMLR}
}

@article{pake,
  title={PAKE: Learning Whole-Body Loco-Manipulation with Partial Kinematic Embeddings},
  author={He, Zhengmao and Jung, Moonkyu and Kim, Hyeongjun and Lee, Jiseong and Zhang, Hui and Hwangbo, Jemin and Song, Jie},
  journal={arXiv preprint arXiv:2607.11041},
  year={2026}
}

@inproceedings{rma,
  title={Rma: Rapid motor adaptation for legged robots},
  author={Kumar, Ashish and Fu, Zipeng and Pathak, Deepak and Malik, Jitendra},
  booktitle = {Robotics: Science and Systems},
  year={2021}
}

@inproceedings{metamorph,
  title={Metamorph: Learning universal controllers with transformers},
  author={Gupta, Agrim and Fan, Linxi and Ganguli, Surya and Fei-Fei, Li},
  booktitle = {International Conference on Learning Representations},
  year={2022}
}

@inproceedings{act,
  title={Learning fine-grained bimanual manipulation with low-cost hardware},
  author={Zhao, Tony Z and Kumar, Vikash and Levine, Sergey and Finn, Chelsea},
  booktitle = {Robotics: Science and Systems},
  year={2023}
}

@article{diffusion_policy,
  title={Diffusion policy: Visuomotor policy learning via action diffusion},
  author={Chi, Cheng and Xu, Zhenjia and Feng, Siyuan and Cousineau, Eric and Du, Yilun and Burchfiel, Benjamin and Tedrake, Russ and Song, Shuran},
  journal={The International Journal of Robotics Research},
  volume={44},
  number={10-11},
  pages={1684--1704},
  year={2025},
  publisher={Sage Publications Sage UK: London, England}
}

@inproceedings{umi,
  title={Universal manipulation interface: In-the-wild robot teaching without in-the-wild robots},
  author={Chi, Cheng and Xu, Zhenjia and Pan, Chuer and Cousineau, Eric and Burchfiel, Benjamin and Feng, Siyuan and Tedrake, Russ and Song, Shuran},
  booktitle = {Robotics: Science and Systems},
  year      = {2024},
}

@inproceedings{rt2,
  title={Rt-2: Vision-language-action models transfer web knowledge to robotic control},
  author={Brohan, Anthony and Brown, Noah and Carbajal, Justice and Chebotar, Yevgen and Chen, Xi and Choromanski, Krzysztof and Ding, Tianli and Driess, Danny and Dubey, Avinava and Finn, Chelsea and others},
  booktitle = {Conference on Robot Learning},
  pages     = {2165--2183},
  volume    = {229},
  year={2023},
  organization={PMLR}
}

@inproceedings{octo,
  title={Octo: An open-source generalist robot policy},
  author={Team, Octo Model and Ghosh, Dibya and Walke, Homer and Pertsch, Karl and Black, Kevin and Mees, Oier and Dasari, Sudeep and Hejna, Joey and Kreiman, Tobias and Xu, Charles and others},
  booktitle = {Robotics: Science and Systems},
  year={2024}
}

@inproceedings{openvla,
  title={Openvla: An open-source vision-language-action model, 2024},
  author={Kim, Moo Jin and Pertsch, Karl and Karamcheti, Siddharth and Xiao, Ted and Balakrishna, Ashwin and Nair, Suraj and Rafailov, Rafael and Foster, Ethan and Lam, Grace and Sanketi, Pannag and others},
  booktitle = {Conference on Robot Learning},
  pages     = {2679--2713},
  volume    = {270},
  year={2024},
  organization={PMLR}
}

@inproceedings{pi0,
  title={{$\pi_0$}: A Vision-Language-Action Flow Model for General Robot Control},
  author={Black, Kevin and Brown, Noah and Driess, Danny and Esmail, Adnan and Equi, Michael and Finn, Chelsea and Fusai, Niccolo and Groom, Lachy and Hausman, Karol and Ichter, Brian and others},
  booktitle = {Robotics: Science and Systems},
  year      = {2025},
}

@article{hybrid_loco_manipulation,
  title={Combining learning-based locomotion policy with model-based manipulation for legged mobile manipulators},
  author={Ma, Yuntao and Farshidian, Farbod and Miki, Takahiro and Lee, Joonho and Hutter, Marco},
  journal={IEEE Robotics and Automation Letters},
  volume={7},
  number={2},
  pages={2377--2384},
  year={2022},
  publisher={IEEE}
}

@inproceedings{modumorph,
  title={Universal morphology control via contextual modulation},
  author={Xiong, Zheng and Beck, Jacob and Whiteson, Shimon},
  booktitle={International Conference on Machine Learning},
  pages={38286--38300},
  year={2023},
  organization={PMLR}
}

@article{mlm,
  title={Mlm: Learning multi-task loco-manipulation whole-body control for quadruped robot with arm},
  author={Liu, Xin and Ma, Bida and Qi, Chenkun and Ding, Yan and Xu, Nuo and Zhang, Guorong and Chen, Pengan and Liu, Kehui and Jia, Zhongjie and Guan, Chuyue and others},
  journal={IEEE Robotics and Automation Letters},
  volume={11},
  number={1},
  pages={81--88},
  year={2025},
  publisher={IEEE}
}

@article{tactile_aware_loco_manipulation,
  title={Learning Tactile-Aware Quadrupedal Loco-Manipulation Policies},
  author={Zhou, Pokuang and Zhou, Yuhao and Luu, Quan Khanh and Han, Seungho and Zhang, Heng and Huang, Binghao and Li, Yunzhu and Ajoudani, Arash and Xu, Zhengtong and She, Yu},
  journal={arXiv preprint arXiv:2604.27224},
  year={2026}
}

@article{dreamtrajectory,
  title={DreamTrajectory: Trajectory-Guided Action Generation with World Model Alignment for Mobile Manipulation},
  author={Yang, Zheng and Zhang, Wenjie and Chen, Xiangyu and Song, Wenxuan and Wang, Xianpeng and Kang, Yihang and Chen, Wen and Wang, Lujia and Xu, Renjing and Chu, Xiaowen},
  journal={arXiv preprint arXiv:2608.01381},
  year={2026}
}

@article{cmp,
  title={CMP: Robust Whole-Body Tracking for Loco-Manipulation via Competence Manifold Projection},
  author={Cheng, Ziyang and Wei, Haoyu and Yin, Hang and Xu, Xiuwei and Yu, Bingyao and Zhou, Jie and Lu, Jiwen},
  journal={arXiv preprint arXiv:2604.07457},
  year={2026}
}

@article{isaaclab,
  title={Isaac lab: A gpu-accelerated simulation framework for multi-modal robot learning},
  author={Mittal, Mayank and Roth, Pascal and Tigue, James and Richard, Antoine and Zhang, Octi and Du, Peter and Serrano-Munoz, Antonio and Yao, Xinjie and Zurbr{\"u}gg, Ren{\'e} and Rudin, Nikita and others},
  journal={arXiv preprint arXiv:2511.04831},
  year={2025}
}

@inproceedings{mujoco,
  title={Mujoco: A physics engine for model-based control},
  author={Todorov, Emanuel and Erez, Tom and Tassa, Yuval},
  booktitle={International conference on intelligent robots and systems},
  pages={5026--5033},
  year={2012},
  organization={IEEE}
}

@article{legged_manipulation_review,
  title={Legged robots for object manipulation: A review},
  author={Gong, Yifeng and Sun, Ge and Nair, Aditya and Bidwai, Aditya and CS, Raghuram and Grezmak, John and Sartoretti, Guillaume and Daltorio, Kathryn A},
  journal={Frontiers in Mechanical Engineering},
  volume={9},
  pages={1142421},
  year={2023},
  publisher={Frontiers Media SA}
}

@article{mobile_manipulation_review,
  title={Motion planning for mobile manipulators—a systematic review},
  author={Sandakalum, Thushara and Ang Jr, Marcelo H},
  journal={Machines},
  volume={10},
  number={2},
  pages={97},
  year={2022},
  publisher={MDPI}
}

@article{operational_space,
  title={A unified approach for motion and force control of robot manipulators: The operational space formulation},
  author={Khatib, Oussama},
  journal={IEEE Journal on Robotics and Automation},
  volume={3},
  number={1},
  pages={43--53},
  year={1987},
  publisher={IEEE}
}

@article{sentis_wbc,
  title={Synthesis of whole-body behaviors through hierarchical control of behavioral primitives},
  author={Sentis, Luis and Khatib, Oussama},
  journal={International Journal of Humanoid Robotics},
  volume={2},
  number={04},
  pages={505--518},
  year={2005},
  publisher={World Scientific}
}

@inproceedings{stormmap,
  title={Generating continuous motion and force plans in real-time for legged mobile manipulation},
  author={Ewen, Parker and Sleiman, Jean-Pierre and Chen, Yuxin and Lu, Wei-Chun and Hutter, Marco and Vasudevan, Ram},
  booktitle={2021 IEEE international conference on robotics and automation},
  pages={4933--4939},
  year={2021},
  organization={IEEE}
}

@inproceedings{collision_free_wb_mpc,
  title={A collision-free mpc for whole-body dynamic locomotion and manipulation},
  author={Chiu, Jia-Ruei and Sleiman, Jean-Pierre and Mittal, Mayank and Farshidian, Farbod and Hutter, Marco},
  booktitle={2022 international conference on robotics and automation},
  pages={4686--4693},
  year={2022},
  organization={IEEE}
}

@article{reachability_priors,
  title={Robot learning of mobile manipulation with reachability behavior priors},
  author={Jauhri, Snehal and Peters, Jan and Chalvatzaki, Georgia},
  journal={IEEE Robotics and Automation Letters},
  volume={7},
  number={3},
  pages={8399--8406},
  year={2022},
  publisher={IEEE}
}

@inproceedings{disturbance_predictive_loco_manipulation,
  title={A transferable legged mobile manipulation framework based on disturbance predictive control},
  author={Yao, Qingfeng and Wang, Jilong and Yang, Shuyu and Wang, Cong and Meng, Linghan and Zhang, Qifeng and Wang, Donglin},
  booktitle={2022 IEEE International Conference on Robotics and Biomimetics},
  pages={2093--2097},
  year={2022},
  organization={IEEE}
}

@inproceedings{nervenet,
  title={NerveNet: Learning Structured Policy with Graph Neural Networks.},
  author={Wang, Tingwu and Liao, Renjie and Ba, Jimmy and Fidler, Sanja},
  booktitle={International Conference on Learning Representations},
  year={2018}
}

@inproceedings{shared_modular_policies,
  title={One policy to control them all: Shared modular policies for agent-agnostic control},
  author={Huang, Wenlong and Mordatch, Igor and Pathak, Deepak},
  booktitle={International Conference on Machine Learning},
  pages={4455--4464},
  year={2020},
  organization={PMLR}
}

@inproceedings{mobile_aloha,
  title={Mobile aloha: Learning bimanual mobile manipulation using low-cost whole-body teleoperation},
  author={Fu, Zipeng and Zhao, Tony Z and Finn, Chelsea},
  booktitle={Conference on Robot Learning},
  year={2024}
}

@inproceedings{locoformer,
  title={Locoformer: Generalist locomotion via long-context adaptation},
  author={Liu, Min and Pathak, Deepak and Agarwal, Ananye},
  booktitle={Conference on Robot Learning},
  year={2025},
  organization={PMLR}
}

@inproceedings{genloco,
  title={Genloco: Generalized locomotion controllers for quadrupedal robots},
  author={Feng, Gilbert and Zhang, Hongbo and Li, Zhongyu and Peng, Xue Bin and Basireddy, Bhuvan and Yue, Linzhu and Song, Zhitao and Yang, Lizhi and Liu, Yunhui and Sreenath, Koushil and others},
  booktitle={Conference on Robot Learning},
  pages={1893--1903},
  year={2023},
  organization={PMLR}
}

@article{ppo,
  title={Proximal policy optimization algorithms},
  author={Schulman, John and Wolski, Filip and Dhariwal, Prafulla and Radford, Alec and Klimov, Oleg},
  journal={arXiv preprint arXiv:1707.06347},
  year={2017}
}

@inproceedings{stdarm,
  title={STDArm: Transferring visuomotor policies from static data training to dynamic robot manipulation},
  author={Duan, Yifan and Li, Heng and Wu, Yilong and Yu, Wenhao and Zhang, Xinran and Shen, Yedong and Ji, Jianmin and Zhang, Yanyong},
  booktitle = {Robotics: Science and Systems},
  year = {2025},
}

@article{cacvla,
  title={CAC-VLA: Context-Gated Action Conditioning for Vision-Language-Action Models},
  author={Xiong, Yifu and Yu, Wenhao and Lin, Jiaxuan and Zou, Bojun and Li, Jiahao and Zhang, Lu and Zhang, Yanyong and Ji, Jianmin},
  journal={arXiv preprint arXiv:2607.04816},
  year={2026}
}

@article{favla,
  title={Favla: A force-adaptive fast-slow vla model for contact-rich robotic manipulation},
  author={Li, Yao and Tang, Peiyuan and Zhang, Wuyang and Zhu, Chengyang and Duan, Yifan and Shi, Weikai and Zhang, Xiaodong and Yang, Zijiang and Ji, Jianmin and Zhang, Yanyong},
  journal={arXiv preprint arXiv:2602.23648},
  year={2026}
}

@article{driftingvla,
  title={DriftingVLA: Native One-Step Vision-Language-Action Generation via Per-Dimension Temporal Drifting},
  author={Gao, Yuxuan and Zhang, Shiqi and Shen, Yedong and Duan, Yifan and Yu, Wenhao and Zhang, Xin and Cao, Siyuan and Deng, Jiajun and Zhang, Yanyong},
  journal={arXiv preprint arXiv:2608.29749},
  year={2026}
}

\end{document}